\pdfoutput=1
\documentclass[10pt]{article}
\usepackage[final]{peace}
\usepackage{tikz}
\usetikzlibrary{arrows.meta,positioning,decorations.pathreplacing}
\newcommand{\EoneRho}{0.146}
\newcommand{\EoneInvRho}{6.8}
\newcommand{\EoneCzero}{0.999}
\newcommand{\EoneCzeroThirty}{0.996}
\newcommand{\EoneCzeroDev}{0.014}
\newcommand{\EoneCachro}{0.999}
\newcommand{\EoneCmulti}{0.765}
\newcommand{\EoneCwrong}{0.985}
\newcommand{\EoneLinZero}{0.985}
\newcommand{\EoneLinFifteen}{0.915}
\newcommand{\EoneLinThirty}{0.657}
\newcommand{\EoneDofpLinear}{1.97}
\newcommand{\EoneDofpAxisOne}{3.4\%}
\newcommand{\EoneDofpAxisThree}{10.3\%}
\newcommand{\EoneLeakRelErr}{10\%}

\newcommand{\EtwoEngNineThr}{5.0}
\newcommand{\EtwoEngEighteenThr}{5.0}
\newcommand{\EtwoKorNineThr}{8.0}
\newcommand{\EtwoKorEighteenThr}{5.0}
\newcommand{\EtwoEngNineVideo}{1.2\times10^{-3}}
\newcommand{\EtwoEngNineSingle}{0.050}

\newcommand{\EtwoKorNineVideo}{1.9\times10^{-3}}

\newcommand{\EtwoEngNinePlain}{1.00}
\newcommand{\EtwoEngNineKeyed}{1.00}

\newcommand{\EtwoKorNinePlain}{0.98}
\newcommand{\EtwoKorNineKeyed}{0.98}

\newcommand{\EthreeReveal}{0.90}
\newcommand{\EthreeFilter}{500}
\newcommand{\EthreeStdDE}{1.7\times10^{-13}}
\newcommand{\EthreeObsMedDE}{9.1}
\newcommand{\EthreeObsPninetyfiveDE}{18.5}
\newcommand{\EthreeObsFracGtOne}{100\%}
\newcommand{\EthreeCamMed}{0.25}
\newcommand{\EthreeCamMin}{0.18}
\newcommand{\EthreeCamMax}{0.71}
\newcommand{\EthreeNcam}{28}
\newcommand{\EthreeNobs}{151}

\title{PEACE-TEST-PROTOCOL: Polarization-Keyed Examination Papers and Mandatory Issued Eyewear that Displace AI Glasses and Extend Universal Design}
\peacekicker{Framework, analysis, simulation study, and evaluation protocol \textperiodcentered\ v1, 21 November 2026}
\peacetitlemark{}
\peacesubtitle{The examination page you can read only through the glasses we hand you}
\peaceauthor{Dohyeon Raphael KWON}
\peaceaffil{PEACE-UNIVERSE, Cheongju, Republic of Korea\\\url{https://peace.ac} \textperiodcentered\ \href{mailto:raphael.kwon@peace.ac}{\texttt{raphael.kwon@peace.ac}}}
\peaceshorttitle{PEACE-TEST-PROTOCOL: keyed papers, issued eyewear}
\peacepreprintid{Preprint}
\begin{document}
\peacemaketitle
\begin{abstract}
Camera-equipped AI glasses turn an examination page into a query for a remote vision--language model. Existing responses detect the device, or (as in our companion work SEOYUN) try to make a captured page useless to the model. PEACE-TEST-PROTOCOL takes a third route. It changes \emph{who owns the eyewear}. Every candidate wears glasses issued by the examination authority for the whole session, and the question paper is printed so that its content is legible only through those glasses. The mechanism is the one cinemas use for stereoscopic film. A polarization-patterned laminate encodes the text as a difference in polarization state rather than in brightness, so the page looks like a uniform grey field to the naked eye and to an unfiltered camera, and resolves into text through an issued circular analyzer. Making issued eyewear compulsory turns the smart-glasses form factor against itself. The face position, from which a covert camera shares the candidate's line of sight, is occupied by a device the authority supplies, inspects, and collects. A camera elsewhere must carry its own matching analyzer to read anything. Candidates who already wear spectacles receive prescription inserts in the issued frame, following the insert model that consumer smart glasses use for prescription lenses. Low-vision candidates may use an authority-owned, offline assistive tier that adds magnification, contrast remapping, and speech. Blind candidates keep braille, audio, and screen-reader formats, so the optical key is never their only route to the questions. We derive the luminance and contrast budget of the keyed page using Malus's law, show why circular rather than linear keying is required for head-tilt tolerance, and explain why spectral (metameric) keying can make content \emph{more} visible to cameras than to eyes. A simulation study supports these analyses. A spectral Mueller model confirms that the circular key is tilt-invariant and invisible to linear-polarization cameras when it is ideally fabricated. A misregistration of only $1^\circ$ between glyph and background axes, however, already leaks a few percent of linear-polarization contrast. A capture-and-OCR experiment on English and Korean text shows that readability by an unfiltered camera depends on the contrast-to-noise ratio, so residual leakage that a single frame misses can be recovered by averaging video. Leakage must therefore be bounded against the adversary's whole photon budget. Using measured colour-matching functions of 151 observers and sensitivities of 28 cameras, we show that a spectral key designed for the standard observer is visible to every real observer and every camera, and that hardening it against known cameras erases the revealed contrast. We map the protocol onto the Universal Design for Learning guidelines and enumerate its failure modes. The key carries no secrecy, analyzers are commodity parts, other covert cameras remain possible, and illumination must be raised. We specify a staged evaluation of optical benches, filtered-camera red-teaming, and human legibility and accessibility studies. All results are simulations. This version reports no physical measurements, prototypes, or human-subject results. The protocol raises the cost of unauthorized capture; it does not make capture impossible, and it is not a basis for allegations against individuals.
\end{abstract}

\section{Introduction}
\label{sec:intro}
In a stereoscopic cinema the screen carries two images at once. Without glasses the viewer sees a blur of both, and with the issued glasses each eye receives only its own image. Nobody in the audience finds the arrangement strange, because the glasses are handed out at the door and returned at the exit. This paper asks whether an examination can be organized the same way. The question paper is printed so that only the issued glasses can read it, and every candidate wears those glasses for the whole session.

The motivation is the arrival of camera-equipped AI glasses in examination rooms. Korean reporting describes candidates caught using AI glasses in national qualification and language examinations, and it questions whether device bans suffice for the College Scholastic Ability Test (CSAT)~\cite{news_daum_qualification,news_tf_csat_ban}. The detailed implementation plan for the 2027 CSAT explicitly lists smart glasses among banned devices~\cite{kice2026csatplan}. These sources motivate a threat model. They are not measurements of how often cheating occurs or how often it succeeds. Current responses fall into two families. \emph{Possession controls}, such as bans, inspections, and radio-frequency surveys~\cite{examrfguard}, ask whether a prohibited device is present. \emph{Content controls}, including our companion framework SEOYUN~\cite{kwon2026seoyun} and defensive visual prompt injection~\cite{shao2026images}, ask whether a captured page is useful to the model that receives it. SEOYUN operates downstream of the camera, on the model's policy and encoder. Whatever it embeds must still be readable by a human, and therefore is also readable by any camera that a human can out-resolve.

PEACE-TEST-PROTOCOL (henceforth PTP) works upstream of the model, on the optical channel and on the eyewear itself. It has three parts.

\begin{enumerate}
\item \textbf{Keyed pages.} Question content is encoded in the polarization state of light leaving the page rather than in its brightness or colour. To the naked eye and to an unfiltered camera the page is a nearly uniform grey field. Through a matching circular analyzer the text appears.
\item \textbf{Mandatory issued eyewear.} Every candidate wears analyzer glasses issued and collected by the authority. Because the eyewear is compulsory and supplied, the face, the one body position from which a covert camera naturally shares the candidate's line of sight, is occupied by a trusted device. We describe this as turning the smart-glasses form factor against itself. The authority does not hunt for a glasses camera; it makes the glasses its own.
\item \textbf{Universal design.} Candidates with spectacles receive prescription inserts in the issued frame, as consumer smart glasses already accept prescription lenses~\cite{meta_rayban_rx}. Low-vision candidates may opt into an authority-owned, offline mixed-reality assistive tier. Blind candidates keep non-optical formats. The protocol is designed against the Universal Design for Learning (UDL) guidelines~\cite{cast2024udl} rather than adding accommodations afterwards.
\end{enumerate}

\paragraph{What PTP does and does not claim}
Polarization keying is not encryption. Analyzers are inexpensive commodity parts, and we show in \S\ref{sec:optics} that a binary circular key has no secrecy at all, since the opposite analyzer shows the same text with inverted polarity. PTP's security claim is narrower. An \emph{unfiltered} camera gets a near-blank page. The most convenient filtered camera position, the face, is taken by the authority's eyewear. Any other capture therefore needs both a concealed camera and a concealed, correctly oriented analyzer, while existing possession controls continue to apply. PTP raises the cost of capture; it does not make capture impossible.

\paragraph{Contributions}
(i) A threat model that separates the \emph{capture channel} from possession and from model behaviour, and that places PTP relative to possession controls and SEOYUN (\S\ref{sec:threat}). (ii) An optical analysis of keyed pages, covering the luminance budget, a contrast model with extinction and retardance error, the head-tilt argument for circular over linear keying, and the camera-metamerism argument against spectral keying (\S\ref{sec:optics}). (iii) An eyewear specification with passive, prescription, and assistive tiers, plus inspection and custody rules (\S\ref{sec:eyewear}). (iv) A UDL mapping and accessibility constraints, including the requirement that no candidate's access to the questions depends solely on the optical key (\S\ref{sec:udl}). (v) A simulation study (\S\ref{sec:sim}) with three experiments: a spectral Mueller model of the keyed page; a capture-and-OCR experiment that turns leakage into a photon-budget requirement; and a test of metameric keys against 151 measured observers and 28 measured cameras. (vi) A failure-mode analysis (\S\ref{sec:security}) and a staged evaluation protocol with prespecified metrics (\S\ref{sec:protocol}).

\paragraph{Scope}
No prototype has been built and no physical measurement has been taken. The quantities in \S\ref{sec:optics} are symbolic or illustrative, and those in \S\ref{sec:sim} come from simulations; both are labelled as such. No examination authority, manufacturer, or provider has endorsed the protocol.

\section{Threat Model}
\label{sec:threat}
\subsection{Pipeline and stages}
An unauthorized assistance pipeline has four stages. (S1)~\emph{Possession}: a capture device is brought into the room. (S2)~\emph{Capture}: the device records the page. (S3)~\emph{Exfiltration}: the image or transcription leaves the device. (S4)~\emph{Solution}: a model or person returns an answer, which reaches the candidate. Possession controls act on S1 and partly S3, radio surveys act on S3~\cite{examrfguard}, and SEOYUN acts on S4~\cite{kwon2026seoyun}. PTP acts on S2. These layers are complementary. Because they fail under different conditions, combining them is meaningful.

\subsection{Adversary}
The adversary is a candidate, possibly helped by an outside party, who wants to capture question content during the session. The adversary may own consumer AI glasses, a concealed body-worn camera, or a polarization-sensitive camera module, and may know the PTP specification in full (Kerckhoffs's principle). The adversary cannot alter the issued eyewear without leaving tamper evidence (\S\ref{sec:eyewear}) and cannot change room lighting or page manufacture. We do not consider insiders who leak papers before the session; PTP leaves that risk unchanged.

\subsection{Capture positions}
\label{sec:positions}
We group capture positions by how well they see the page:
\begin{itemize}
\item \textbf{Face (F).} A camera in a spectacle frame shares the candidate's line of sight, follows head motion while reading, and looks like ordinary eyewear. This is the position AI glasses exploit.
\item \textbf{Body (B).} Buttons, collars, wristbands, pens. These need deliberate aiming and produce oblique, partly occluded views.
\item \textbf{Desk (D).} Objects placed on the desk, such as calculators or erasers. These are constrained by the rules on desk contents and are visible to proctors.
\item \textbf{Room (R).} Cameras outside the candidate's control. Out of scope.
\end{itemize}
PTP makes two changes. Position F is assigned to the authority's device. Positions B and D see a near-blank page unless the camera carries an analyzer of the right type, aligned well enough to give usable contrast. \S\ref{sec:security} discusses how much this raises the adversary's cost.

\subsection{Defender goals and constraints}
\begin{enumerate}
\item[G1] Unfiltered captures must contain no recoverable question content, as measured by optical character recognition (OCR) and vision--language model extraction (\S\ref{sec:protocol}).
\item[G2] Candidates wearing the issued eyewear must read keyed pages with accuracy, speed, and comfort not inferior to plain paper, within a margin fixed in advance.
\item[G3] No candidate may be disadvantaged by a visual impairment, a refractive error, or a disability, and no candidate's only access to the questions may run through the optical key.
\item[G4] The eyewear must not collect personal data, and the protocol must not produce suspicion scores for individuals.
\end{enumerate}

\section{Related Work}
\label{sec:related}
\paragraph{Stereoscopic projection}
Cinema stereoscopy multiplexes two images by polarization or by wavelength. Polarization-switched stereoscopic displays adopted circular rather than linear polarization because, with linear states, ``head-tipping of only a few degrees'' raises crosstalk, whereas with circular states it does not~\cite{lipton1987circular,lipton1988pushpull}. We rely on this property in \S\ref{sec:tilt} and quantify it in \S\ref{sec:sim}. Wavelength-multiplexing systems use interference filters with interleaved narrow passbands~\cite{jorke2006stereo}. PTP borrows the viewing arrangement but has a different goal. Cinema glasses separate two images that are both meant to be seen. PTP glasses reveal one image that is meant to stay hidden from viewers without them.

\paragraph{Polarization optics and patterned retarders}
Malus's law, wave plates, and the generation of circular polarization are standard~\cite{hecht2017optics,born1999principles}. Cholesteric liquid crystals selectively reflect one handedness of circular polarization~\cite{degennes1993physics}, and photo-alignment allows liquid-crystal retarder films to be patterned at high resolution~\cite{chigrinov2008photoalignment}. Film-patterned retarders are produced at display scale, although conventional processes rely on photo-alignment through fixed masks~\cite{lim2014fpr}. Photo-patterned liquid-crystal-polymer images that are ``invisible when viewed with the naked eye but becomes visible when viewed through a polarizer'' are an established optical security element~\cite{moia2000lpp}, and cholesteric films combined with patterned quarter-wave plates have been proposed for anti-counterfeiting~\cite{zang2026clc,li2017clc}. Our contribution is not the optical element. It is the analysis of such elements for examinations, covering luminance, tilt, camera leakage, and accessibility, and their combination with compulsory eyewear.

\paragraph{Human polarization sensitivity}
Humans are not entirely polarization-blind. Haidinger's brushes are a faint entoptic pattern seen in polarized light, and human sensitivity to them has been measured~\cite{temple2015perceiving}. Brushes have also been reported under circularly polarized light, attributed to corneal retardance that varies widely between people~\cite{shurcliff1955haidinger,devries1950properties,temple2015perceiving}. A circularly keyed page is therefore not guaranteed to be invisible to every unaided observer. Whether such cues could ever convey text is untested and must be checked in the human study (\S\ref{sec:protocol}).

\paragraph{Visual cryptography}
Naor and Shamir showed that a secret image can be split into two printed shares that reveal the image only when overlaid~\cite{naor1995visual}. This provides real secrecy, but it requires pixel-accurate registration between the shares. A lens held at reading distance cannot keep that registration, so we use it only for comparison.

\paragraph{Metamerism}
Two spectra that match for one observer generally do not match for another~\cite{wyszecki1982color}. Colour-matching functions vary across colour-normal people~\cite{asano2016individual}, and camera sensitivities satisfy the Luther condition only approximately~\cite{jiang2013space}. In \S\ref{sec:spectral} and \S\ref{sec:sim} we show with measured observer and camera data that this observer dependence defeats spectrally keyed pages.

\paragraph{AI glasses and examinations}
Physical prompt injection against wearable vision--language systems has been analysed directly~\cite{li2026lens}. Content-level exam defences seek model refusal~\cite{kwon2026seoyun}, building on image perturbations shown to induce refusal in other settings~\cite{shao2024refusing,shao2026images}. Radio surveys compare Wi-Fi and Bluetooth signals with an empty-room baseline, without claiming to detect every device~\cite{examrfguard}. Division-of-focal-plane polarization sensors use on-chip linear wire-grid analyzers~\cite{maruyama2018polarization,york2014bioinspired}, which is relevant to whether a polarization camera can read a circular key (\S\ref{sec:camleak}). We know of no prior proposal that makes the authority's eyewear compulsory as a means of excluding personal eyewear.

\paragraph{Head-mounted low-vision aids and universal design}
Head-mounted systems that enhance a low-vision user's view, by magnification, contrast enhancement, or cues, have been designed and evaluated~\cite{zhao2015foresee,zhao2016cuesee}. Consumer smart glasses now come with prescription lenses, from $-6.00$ to $+6.00$ on several current models~\cite{meta_rayban_rx}, and offer camera-based scene description and volunteer video assistance for blind and low-vision users~\cite{meta2026accessibility,bemyeyes2026meta}. UDL asks designers to \emph{design multiple means} of engagement, representation, and action and expression~\cite{rose2002teaching,cast2024udl}. Universal design for large-scale assessment applies these principles to tests~\cite{thompson2002universal}. Article~24 of the Convention on the Rights of Persons with Disabilities~\cite{un2006crpd} and Korea's anti-discrimination statute~\cite{korea2007disability} set the legal baseline that PTP must not lower.

\section{Optics of the Keyed Page}
\label{sec:optics}
We describe the light leaving a page element by its Stokes vector $(S_0,S_1,S_2,S_3)$~\cite{born1999principles}. An ordinary printed page encodes text in $S_0$ (brightness) and in its spectral distribution (colour). A keyed page keeps $S_0$ and the spectrum uniform and encodes text only in $S_3$, the sign of circular polarization: left-handed ($S_3>0$) for the background and right-handed ($S_3<0$) for the glyphs. Any observer that measures only intensity, whether an eye or a camera, receives the same $S_0$ from both and sees a uniform field.

\subsection{Two constructions}
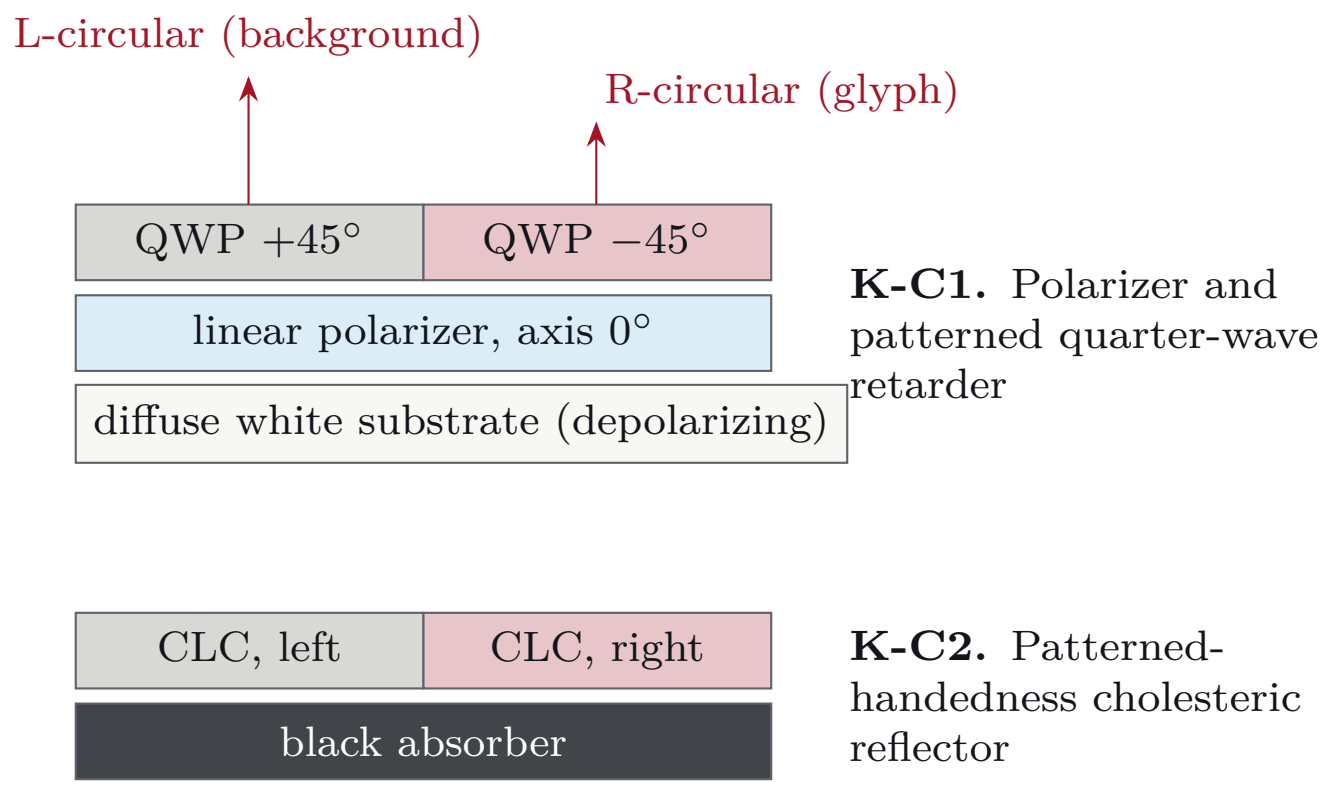
\begin{figure}[t]
\centering
\begin{tikzpicture}[font=\footnotesize,>=Stealth,
 layer/.style={draw=peacemute,minimum width=46mm,minimum height=5mm,anchor=west}]
\node[layer,fill=peacepaper] (sub) at (0,0) {diffuse white substrate (depolarizing)};
\node[layer,fill=peaceblue!25] (pol) at (0,0.6) {linear polarizer, axis $0^\circ$};
\node[layer,minimum width=23mm,fill=peacerule] (qa) at (0,1.2) {QWP $+45^\circ$};
\node[layer,minimum width=23mm,fill=peacedeep!25] (qb) at (2.3,1.2) {QWP $-45^\circ$};
\node[anchor=west,text width=34mm] at (5.0,0.6) {\textbf{K-C1.} Polarizer and patterned quarter-wave retarder};
\draw[->,peacedeep] (1.15,1.45) -- (1.15,2.3) node[above]{L-circular (background)};
\draw[->,peacedeep] (3.45,1.45) -- (3.45,2.0) node[above right=-1mm]{R-circular (glyph)};
\begin{scope}[yshift=-2.1cm]
\node[layer,fill=peaceink!80,text=white] (blk) at (0,0) {black absorber};
\node[layer,minimum width=23mm,fill=peacerule] at (0,0.6) {CLC, left};
\node[layer,minimum width=23mm,fill=peacedeep!25] at (2.3,0.6) {CLC, right};
\node[anchor=west,text width=34mm] at (5.0,0.3) {\textbf{K-C2.} Patterned-handedness cholesteric reflector};
\end{scope}
\end{tikzpicture}
\caption{Two constructions of a keyed page (schematic, not to scale). In both, background and glyph regions return equal intensity in opposite circular states. The issued analyzer passes left-circular light and blocks right-circular light, so the glyphs appear dark.}
\label{fig:stack}
\end{figure}
\textbf{K-C1} (Fig.~\ref{fig:stack}, top) laminates a linear polarizer over a diffuse white substrate and places a patterned quarter-wave retarder (QWP) above it. The fast axis is at $+45^\circ$ in background regions and $-45^\circ$ in glyph regions. Light reflected by the substrate is depolarized, leaves the polarizer linearly polarized, and leaves the retarder left- or right-circular. Photo-aligned liquid-crystal polymers can pattern such retarders at sub-glyph resolution~\cite{chigrinov2008photoalignment}. \textbf{K-C2} (bottom) prints cholesteric liquid crystal (CLC) of opposite handedness in the two regions over a black absorber. Each CLC selectively reflects light of its own handedness within its band~\cite{degennes1993physics}. K-C2 avoids the double-pass polarizer loss but needs a broadened reflection band to look neutral rather than coloured.

\subsection{Luminance budget}
\label{sec:luminance}
Let $T_u$ be the polarizer's single-pass transmittance for unpolarized light (ideal $0.5$), and $T_a$ the issued analyzer's transmittance for the matched circular state (ideal $1$). For K-C1 with substrate reflectance $R$ under illuminance $E$, the Lambertian background luminance seen through the analyzer is
\begin{equation}
L_b = \frac{R E}{\pi}\, T_u^{2}\, T_a ,
\qquad
\rho \equiv \frac{L_b}{L_{\text{plain}}} = T_u^{2}\,T_a ,
\label{eq:rho}
\end{equation}
where $L_{\text{plain}}=RE/\pi$ is plain paper viewed without glasses. Ideal components give $\rho=0.25$. The illustrative values $T_u=0.42$ and $T_a=0.85$ give $\rho\approx0.15$. These are assumptions, not measured parts. Matching plain-paper retinal illuminance would then require raising desk illuminance by about $1/\rho\approx6.7$. From the 300\,lx Korean classroom minimum~\cite{korea_schoolhealth_rule}, that means roughly 2000\,lx. Because the glasses also attenuate the unpolarized surroundings by roughly $T_u$, the page looks about $T_uT_a\approx0.36$ as bright as unkeyed scratch paper beside it, rather than $0.15$ as bright. Even so, luminance is the protocol's largest human-factors cost. It motivates K-C2, reflective-polarizer variants that recycle the rejected state, and a lighting requirement in the protocol (R5, \S\ref{sec:rules}).

\subsection{Contrast with imperfect parts}
Let $\varepsilon$ be the analyzer's transmittance for the opposite circular state, and let the page retarder deviate from quarter-wave by a retardance error $\delta$. The error leaves a fraction $\sin^2(\delta/2)$ of the light in the opposite state. Through the analyzer,
\begin{align}
L_b &\propto T_a\cos^2\tfrac{\delta}{2} + \varepsilon\sin^2\tfrac{\delta}{2}, &
L_g &\propto \varepsilon\cos^2\tfrac{\delta}{2} + T_a\sin^2\tfrac{\delta}{2},
\end{align}
and the Weber contrast of a glyph is $C = 1 - L_g/L_b$. Neglecting dispersion of birefringence, a zero-order plate designed for $\lambda_0=550$\,nm has $\delta(\lambda)=\tfrac{\pi}{2}(\lambda_0/\lambda-1)$. That gives $\sin^2(\delta/2)\approx0.030$ at 450\,nm and $\approx0.015$ at 650\,nm. Even a non-achromatic retarder therefore keeps crosstalk at a few percent across the visible band, and glyph contrast is limited mainly by $\varepsilon/T_a$. These are calculations from the stated model, not measurements.

\subsection{Why circular keying: head tilt}
\label{sec:tilt}
A linear key would mark glyphs with polarization rotated by $90^\circ$ relative to the background. An analyzer tilted by $\theta$ about the line of sight, as when a candidate tilts their head, gives $L_b\propto\cos^2\theta$ and $L_g\propto\sin^2\theta$, so
\begin{equation}
C_{\text{lin}}(\theta) = 1-\tan^2\theta .
\end{equation}
Contrast falls to about $0.93$ at $15^\circ$ and $0.67$ at $30^\circ$, vanishes at $45^\circ$, and inverts beyond. An ideal circular analyzer is invariant to rotation about the viewing axis, so $C$ does not depend on $\theta$. This is the same reason cinema polarization systems use circular states. Circular keying also matters for the camera argument below.

\subsection{What an unfiltered camera sees}
\label{sec:camleak}
Background and glyphs differ only in the sign of $S_3$, and both have $S_1=S_2=0$. An intensity camera therefore receives equal signal from both. So does a camera with linear polarization sensitivity, including division-of-focal-plane polarization sensors that measure $S_1$ and $S_2$ only, because any linear diattenuator transmits left- and right-circular light equally. Leakage requires incidental retardance $\Gamma$ in the camera path (stress birefringence in plastic optics, cover glass, protective films) followed by a linear diattenuation $D$ (oblique Fresnel surfaces, coatings). A first-order Mueller-calculus estimate gives
\begin{equation}
C_{\text{cam}} \approx 2\,D\,\sin\Gamma ,
\end{equation}
which is small when either factor is small but is not zero. Two other leakage paths are not polarimetric: structural visibility of the patterned layer (thickness steps, edge scattering, raking-light relief) and Haidinger's-brush-type cues for human observers~\cite{temple2015perceiving}. Goal G1 must therefore be tested empirically, with OCR and model extraction on captures from real devices (\S\ref{sec:protocol}). It cannot be assumed from the ideal model, and E2-sim (\S\ref{sec:e2}) shows that the requirement scales with the adversary's photon budget.

\subsection{Why not spectral keying}
\label{sec:spectral}
Wavelength-multiplexed cinema~\cite{jorke2006stereo} suggests another route: print glyphs and background with a metameric pair of inks that match for the human eye but differ inside a narrow band, and reveal the glyphs with a notch filter. The difficulty is that metamerism depends on the observer~\cite{wyszecki1982color}. Camera colour filters do not satisfy the Luther condition, meaning they are not linear combinations of the human colour-matching functions. Measured sensitivities of 28 cameras confirm that the Luther condition holds only approximately~\cite{jiang2013space}, and colour-matching functions also vary between colour-normal people~\cite{asano2016individual}. A pair designed to match for the standard observer will generally \emph{not} match for a camera, nor for a particular person. A spectrally keyed page could therefore be invisible to candidates without glasses and still legible to an unfiltered camera, which inverts the intended asymmetry. A spectral key would have to be metameric simultaneously for the standard observer and for the full population of deployed camera sensitivities. E3-sim (\S\ref{sec:e3}) tests this with measured data. We do not use spectral keying in PTP.

\subsection{Secrecy}
\label{sec:secrecy}
The key carries no secrecy. With a binary circular key, the opposite analyzer shows the same glyphs with inverted polarity. With a linear key, an adversary finds the right angle by rotating a filter. Visual cryptography would provide secrecy~\cite{naor1995visual}, but it needs a second share registered to the page at the pixel level, and spectacles cannot hold that registration. PTP's optical layer therefore achieves G1 against \emph{unfiltered} capture only. Exclusion of filtered capture rests on the eyewear rules in \S\ref{sec:eyewear} and on existing possession controls.

\subsection{Computer-based testing variant}
A liquid-crystal display forms its image by rotating polarization pixel by pixel; the front polarizer converts that rotation into brightness. Remove the front polarizer and add a quarter-wave film at $45^\circ$. Pixel states then leave the display as left- or right-circular light (elliptical light for grey levels), and the analyzing function moves into the issued glasses. Unfiltered observers see a uniform field. This uses existing display stacks, so it may be easier to build than variable-content keyed paper, and it should be evaluated in parallel.

\section{Simulation Study}
\label{sec:sim}
We ran three simulations to test the analysis in \S\ref{sec:optics}. They use measured spectral data where it exists, but every result below concerns \emph{simulated} pages, cameras, and readers. None is a measurement of a physical sample, a real device, or a human participant. Code and the provenance of the data are listed in Appendix~\ref{app:repro}. All numbers in this section are generated from the run logs.

\subsection{E1-sim: spectral Mueller model of the keyed page}
\paragraph{Setup}
Pages and analyzers are modelled as Mueller-matrix chains at 31 wavelengths from 400 to 700\,nm, weighted by CIE illuminant D65 and the CIE 1931 luminosity function $\bar y(\lambda)$. Polarizers have $T_u=0.42$ and polarization efficiency $0.999$. The substrate reflectance is $R=0.85$. Retarders are zero-order quarter-wave plates designed for 550\,nm unless stated otherwise. Glyph contrast is the luminance-weighted Weber contrast.

\paragraph{Results}
\begin{figure}[t]
\centering
\includegraphics[width=\linewidth]{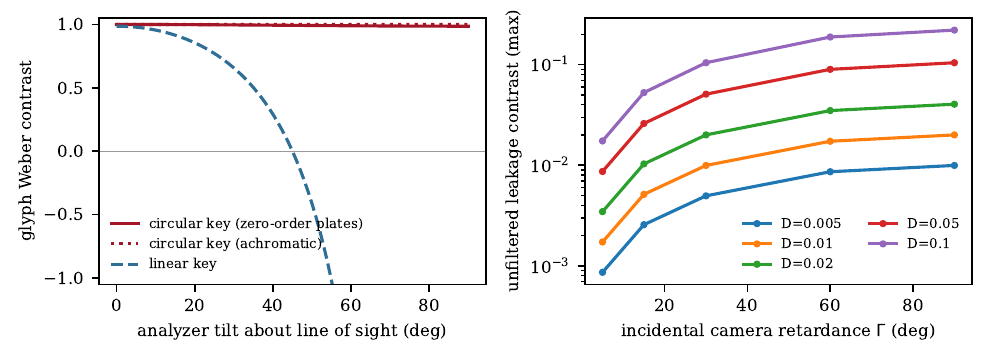}
\caption{E1-sim. Left: glyph contrast through the analyzer as a function of head tilt. The circular key is tilt-invariant; the linear key falls to zero at $45^\circ$ and inverts beyond. Right: worst-case contrast leaked to an unfiltered camera with incidental retardance $\Gamma$ and diattenuation $D$, over random axis orientations.}
\label{fig:exp1}
\end{figure}
\begin{itemize}
\item \emph{Naked eye and intensity cameras.} Background and glyph luminance are identical (contrast $0$) for zero-order, multi-order, and achromatic plates. This holds by construction, because ideal retarders are lossless, so it does not cover structural visibility (\S\ref{sec:camleak}).
\item \emph{Luminance.} $\rho=\EoneRho$, which means the page looks $\EoneInvRho\times$ dimmer than plain paper. Matching the retinal illuminance of plain paper at the 300\,lx Korean classroom minimum~\cite{korea_schoolhealth_rule} would need about $300/\rho\approx2000$\,lx at the desk, which raises its own glare and comfort questions for E2.
\item \emph{Head tilt} (Fig.~\ref{fig:exp1}, left). The circular key gives contrast $\EoneCzero$ at $0^\circ$ and $\EoneCzeroThirty$ at $30^\circ$, varying by at most $\EoneCzeroDev$ over $0$--$90^\circ$. The linear key gives $\EoneLinZero$, $\EoneLinFifteen$, and $\EoneLinThirty$ at $0^\circ$, $15^\circ$, and $30^\circ$, and $0$ at $45^\circ$.
\item \emph{Plate order.} Achromatic plates give $\EoneCachro$. A second-order page retarder lowers contrast to $\EoneCmulti$, so page retarders must be zero-order or achromatic.
\item \emph{Wrong handedness.} Glasses of the opposite handedness show bright glyphs on a dark ground with contrast $\EoneCwrong$. This confirms that the key has no secrecy (\S\ref{sec:secrecy}).
\item \emph{Polarization cameras.} For the linear key, the normalized linear Stokes components differ by $\EoneDofpLinear$ between glyph and background, so a division-of-focal-plane camera reads it directly. For the circular key with ideal axes the difference is zero to numerical precision, and a uniform retardance error does not change this. A misregistration of the glyph fast axis by $1^\circ$ or $3^\circ$, however, produces differences of $\EoneDofpAxisOne$ and $\EoneDofpAxisThree$. By the E2-sim results below, a few percent of contrast is readable after modest frame averaging. \emph{Fabrication tolerance is therefore a security parameter}, and axis registration well below $1^\circ$ is needed.
\item \emph{Leakage formula.} Over the $(D,\Gamma)$ grid, the worst case across random orientations agrees with $2D\sin\Gamma$ (Appendix~\ref{app:leak}) to within $\EoneLeakRelErr$.
\end{itemize}

\subsection{E2-sim: can an unfiltered camera read residual leakage?}
\label{sec:e2}
\paragraph{Setup}
Eight English and eight Korean exam-style lines, written by the author and containing no real exam content, are rendered as glyph masks with Weber contrast $C$. Each image is then passed through a simulated camera: a Gaussian point-spread function, sampling at an x-height of 9 or 18 pixels (the 9-pixel case matches the sampling estimate in~\cite{kwon2026seoyun}), Poisson shot noise at $N_e\in\{10^3,10^4\}$ electrons per pixel, 3\,e$^-$ read noise, and 10-bit quantization. $N_f\in\{1,16,256\}$ registered frames are averaged, which models a static camera recording video. The images are then enhanced as an adversary would, by background flattening, a robust contrast stretch, and $3\times$ upsampling, and read with Tesseract~5.5 (\texttt{eng}, \texttt{kor}). The metric is character accuracy, $1-\text{CER}$, averaged over two noise draws.

\paragraph{Results}
\begin{figure}[t]
\centering
\includegraphics[width=\linewidth]{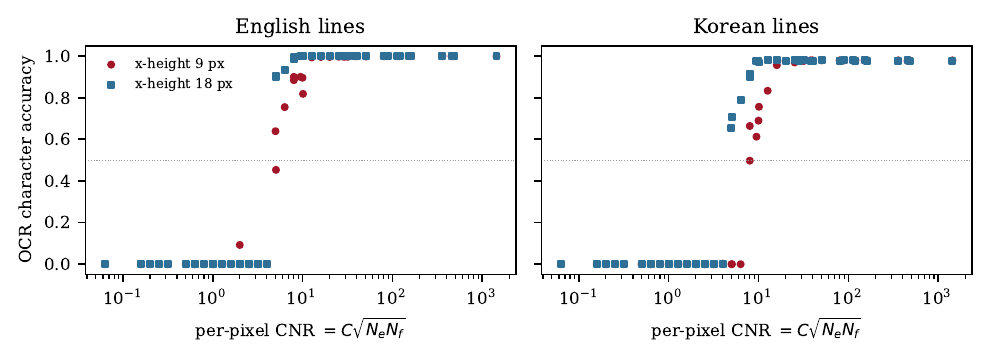}
\caption{E2-sim. OCR character accuracy against the shot-noise-limited per-pixel contrast-to-noise ratio $C\sqrt{N_eN_f}$, for all leakage contrasts, exposures, and frame counts. Accuracy is governed by CNR rather than by $C$ alone.}
\label{fig:exp2}
\end{figure}
\begin{table}[t]
\centering\small
\begin{tabular}{@{}r ccc ccc@{}}
\toprule
& \multicolumn{3}{c}{English, $N_f=$} & \multicolumn{3}{c}{Korean, $N_f=$}\\
\cmidrule(lr){2-4}\cmidrule(l){5-7}
$C$ & 1 & 16 & 256 & 1 & 16 & 256 \\
\midrule
0 & 0.00 & 0.00 & 0.00 & 0.00 & 0.00 & 0.00 \\
0.002 & 0.00 & 0.00 & 0.00 & 0.00 & 0.00 & 0.00 \\
0.005 & 0.00 & 0.09 & 0.90 & 0.00 & 0.00 & 0.50 \\
0.01 & 0.00 & 0.00 & 1.00 & 0.00 & 0.00 & 0.96 \\
0.02 & 0.00 & 0.88 & 1.00 & 0.00 & 0.66 & 0.98 \\
0.05 & 0.64 & 1.00 & 1.00 & 0.00 & 0.98 & 0.98 \\
0.1 & 0.90 & 1.00 & 1.00 & 0.69 & 0.98 & 0.98 \\
0.3 & 1.00 & 1.00 & 1.00 & 0.98 & 0.98 & 0.98 \\
0.9 & 1.00 & 1.00 & 1.00 & 0.98 & 0.98 & 0.98 \\
\bottomrule

\end{tabular}
\caption{E2-sim. OCR character accuracy at x-height 9\,px and $N_e=10^4$, by leakage contrast $C$ and number of averaged frames $N_f$.}
\label{tab:exp2}
\end{table}
Figure~\ref{fig:exp2} shows that OCR success collapses onto a single variable, the per-pixel CNR $C\sqrt{N_eN_f}$. Among the tested conditions, the smallest CNR at which character accuracy reached $0.5$ was $\EtwoEngNineThr$ (English) and $\EtwoKorNineThr$ (Korean) at an x-height of 9\,px, and $\EtwoEngEighteenThr$ and $\EtwoKorEighteenThr$ at 18\,px. Table~\ref{tab:exp2} shows the practical consequence. \emph{Leakage that is unreadable in one frame becomes readable when frames are averaged.} At $N_e=10^4$ a single frame needs $C\gtrsim\EtwoEngNineSingle$ for English. A static camera that averages 60\,s of 30\,fps video ($N_f=1800$) reaches the same CNR at $C\approx\EtwoEngNineVideo$ (English) or $\EtwoKorNineVideo$ (Korean). Goal G1 therefore cannot be stated as ``leakage is small''. It must be stated against the adversary's total photon budget:
\begin{equation}
C_{\text{cam}}\sqrt{N_eN_f} < \text{CNR}^{*},
\end{equation}
where $\text{CNR}^{*}$ is the reading threshold of the \emph{best} available reader. Tesseract is a weak reader at low contrast, and a vision--language model or a dedicated deep reader would probably succeed at lower CNR. The thresholds reported here are therefore optimistic for the defender. We also passed the analyzer view to the same reader as a crude legibility proxy, combining the contrast from E1-sim with exposure reduced by $\rho$. Character accuracy was $\EtwoEngNineKeyed$ (keyed) against $\EtwoEngNinePlain$ (plain) for English and $\EtwoKorNineKeyed$ against $\EtwoKorNinePlain$ for Korean at an x-height of 9\,px. The luminance loss does not limit a machine reader at these exposures, but this says nothing about human comfort, which E2 must measure.

\subsection{E3-sim: spectral keying against real observers and cameras}
\label{sec:e3}
\paragraph{Setup}
We designed ink-reflectance pairs $r_0\pm d/2$ with $r_0=0.5$ by linear programming. The difference spectrum $d$ is constrained to be metameric for the CIE 1931 standard observer under D65, bounded by $|d_i|\le0.45$, and limited in curvature to $|\Delta^2 d_i|\le0.12$ per 10\,nm. Within these constraints the program maximizes the contrast revealed through a 20\,nm band-pass filter, with the best centre chosen by search. The pairs were evaluated on the individual colour-matching functions of \EthreeNobs\ colour-normal observers~\cite{asano2016individual} and on the measured spectral sensitivities of \EthreeNcam\ cameras~\cite{jiang2013space}. Camera contrast is the largest per-channel Weber difference in raw response.

\paragraph{Results}
\begin{figure}[t]
\centering
\includegraphics[width=\linewidth]{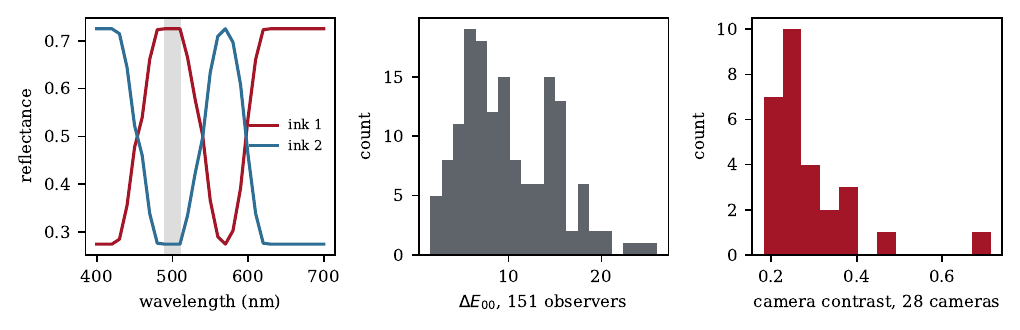}
\caption{E3-sim. Left: an eye-metameric ink pair and its reveal band (grey). Centre: colour difference seen by \EthreeNobs\ individual observers. Right: contrast seen by \EthreeNcam\ cameras.}
\label{fig:exp3}
\end{figure}
The eye-only design reveals text through the filter with contrast $\EthreeReveal$ (centre at \EthreeFilter\,nm), and its difference for the standard observer is $\Delta E_{00}=\EthreeStdDE$. Real observers disagree with the standard observer (Fig.~\ref{fig:exp3}). The median $\Delta E_{00}$ across individuals is \EthreeObsMedDE, the 95th percentile is \EthreeObsPninetyfiveDE, and \EthreeObsFracGtOne\ of observers exceed $\Delta E_{00}=1$. Every camera sees the text, with contrast between \EthreeCamMin\ and \EthreeCamMax\ (median \EthreeCamMed). By the E2-sim thresholds, that contrast is readable in a single frame. Table~\ref{tab:exp3} shows what happens when the design is hardened by also nulling the responses of $k$ randomly chosen training cameras. Leakage to held-out cameras falls, but the revealed contrast falls with it, and from $k=14$ onward only the trivial solution $d=0$ remains. Spectral keying thus trades reveal against leakage, and it must be designed against every camera an adversary might use. Polarization keying does not face this trade-off, because $S_3$ is invisible to \emph{every} intensity-only observer regardless of its spectral sensitivity.
\begin{table}[t]
\centering\small
\begin{tabular}{@{}rcccc@{}}
\toprule
$k$ & reveal contrast & held-out median & held-out max & observer $\Delta E_{00}$ (p95) \\
\midrule
1 & 0.900 & 0.0378 & 0.2502 & 5.56 \\
3 & 0.480 & 0.0135 & 0.1171 & 0.89 \\
7 & 0.108 & 0.0014 & 0.0138 & 0.15 \\
14 & 0.000 & 0.0000 & 0.0000 & 0.00 \\
21 & 0.000 & 0.0000 & 0.0000 & 0.00 \\
\bottomrule

\end{tabular}
\caption{E3-sim. Hardened spectral keys nulling $k$ training cameras. Values are medians over 20 random camera splits. ``Held-out'' is the camera contrast on the remaining $28-k$ cameras.}
\label{tab:exp3}
\end{table}

\subsection{What the simulations do not show}
They do not model structural visibility of patterned layers, specular glare, real polarizer spectra, camera motion, or registration error. They also do not include vision--language model readers or any human perception. In particular, E1-sim's zero naked-eye contrast is a property of ideal components. The physical leakage $C_{\text{cam}}$ is the quantity that E1 on a bench must measure, against the budget derived from E2-sim.

\section{Issued Eyewear: Turning the Smart-Glasses Form Factor Against Itself}
\label{sec:eyewear}
AI glasses work because spectacles are socially invisible. Millions of people wear them, nobody asks why, and they sit exactly where a camera sees what the wearer sees. PTP keeps this property and transfers ownership: the face position (F, \S\ref{sec:positions}) becomes the authority's. Because the keyed page is unreadable without the analyzer, wearing the issued eyewear is not an extra rule that candidates are asked to obey. It is a precondition for taking the examination, and every candidate has a direct reason to comply.

\subsection{Tiers}
\label{sec:tiers}
\begin{table}[t]
\centering\small
\begin{tabularx}{\linewidth}{@{}lXXl@{}}
\toprule
Tier & Who & Device & Electronics \\
\midrule
0 & All candidates by default & Circular-analyzer lenses in a transparent, serialized, tamper-evident frame & None \\
0-Rx & Candidates who wear spectacles & Tier-0 frame with a prescription insert behind the analyzer, or an inspected fit-over variant & None \\
A & Low-vision candidates, by choice & Authority-owned offline mixed-reality eyewear with camera behind the analyzer & Yes, attested \\
B & Blind candidates, or any candidate for whom optics are unsuitable & No optical key: braille, audio, screen reader, large print in a supervised room & Authority devices only \\
\bottomrule
\end{tabularx}
\caption{Eyewear tiers. Tier choice belongs to the candidate, within documented eligibility. No tier is a penalty or a sign of suspicion.}
\label{tab:tiers}
\end{table}

\paragraph{Tier 0: passive key glasses}
Tier-0 glasses are cinema glasses built to examination standards. They have circular-analyzer lenses and no electronics, and the frame is made of transparent material so that any embedded component would be visible. Each frame carries a serial number and a tamper-evident seal. Glasses are issued at entry, recorded against the seat, and collected and reconciled at exit. Hygiene is handled by single use or by validated sanitization between sessions. Because the frame contains no electronics, it cannot record, transmit, or store anything, which satisfies G4 by construction.

\paragraph{Tier 0-Rx: prescription}
Many candidates need refractive correction. The protocol must not force them to choose between their prescription and the key, and it must not give them a loophole by letting personal spectacles, which may hide a camera, stay on the face. Consumer smart glasses solve the same problem by building prescription lenses into the device, at $-6.00$ to $+6.00$ on several current models~\cite{meta_rayban_rx}. PTP adopts that model in reverse. The \emph{authority's} frame accepts a prescription insert. Three routes are specified, in order of preference:
\begin{enumerate}
\item \emph{Registered prescription.} At registration, as with existing accommodation applications, the candidate submits a prescription and receives a sealed Tier-0-Rx pair made by a partner optician, collected on examination day.
\item \emph{Standard insert carrier.} The issued frame has a clip-in carrier behind the analyzer. Partner opticians cut candidates' lenses to the carrier's standard shape in advance. Lenses alone cannot hide a camera, so the carrier is inspected and the frame remains the authority's.
\item \emph{Inspected fit-over.} A fit-over Tier-0 frame worn over the candidate's own passive spectacles, admitted only after inspection and a radio survey~\cite{examrfguard}. This route keeps a residual risk, because the inner frame is personal, and it is reserved for candidates who could not use routes 1 or 2. It is recorded as such.
\end{enumerate}
Candidates who wear contact lenses use Tier 0 directly.

\paragraph{Tier A: assistive mixed reality}
Some low-vision candidates gain from a keyed page read through the analyzer with magnification, contrast remapping, polarity inversion, edge enhancement, or a reading guide. These are the functions of head-mounted low-vision aids~\cite{zhao2015foresee,zhao2016cuesee}. Some also want selected text read aloud. Tier A is an authority-owned headset whose camera sits \emph{behind} the analyzer, so it sees the keyed page as the candidate would. It runs only an attested, offline function set: display enhancement, on-device text recognition, and speech output. Its radio hardware is physically absent or disabled and verified. It contains no model capable of generating answers, keeps no data beyond the session, and is wiped and re-attested between sessions. Tier A has a camera, but the risk is controlled: the authority owns and audits the device, and a camera that only re-renders the page to the candidate who is already reading it adds no new channel. Tier A is offered, never imposed.

\paragraph{Tier B: non-optical access}
For blind candidates, and for any candidate for whom the keyed page or a head-mounted device is unsuitable (for example because of photophobia, nystagmus, or sensory sensitivity), access must not depend on the optical key at all. Tier B provides the formats that examination authorities already supply, in individually supervised rooms. For the CSAT these include braille papers, a computer with screen-reader software and the question file, braille notetakers, and extended time of $1.7\times$ or $1.5\times$ depending on severity~\cite{kice2026csatplan}; large print and audio are further options. Tier B is not a degraded fallback. It is how G3 holds: \emph{no candidate's only access to the questions passes through the optical key.}

\subsection{Protocol rules}
\label{sec:rules}
\begin{enumerate}
\item[R1] \textbf{Universal wearing.} Every candidate in a keyed room wears Tier-0 or Tier-0-Rx eyewear for the whole session. Candidates who should not wear it are moved to Tier A or B, not exempted inside the keyed room.
\item[R2] \textbf{No personal eyewear on the face}, except inspected passive spectacles under the fit-over route. Personal smart glasses remain prohibited under existing rules~\cite{kice2026csatplan}.
\item[R3] \textbf{Custody.} Eyewear is serialized, sealed, issued, reconciled, and collected. A broken seal leads to replacement and a neutral record, never an allegation.
\item[R4] \textbf{Keyed materials.} Question booklets are keyed. Answer sheets and scratch paper stay unkeyed so that marking is unchanged.
\item[R5] \textbf{Lighting.} Desk illuminance is raised above the 300\,lx classroom minimum~\cite{korea_schoolhealth_rule} to the level fixed after the human study (\S\ref{sec:luminance}, E2), with glare control.
\item[R6] \textbf{Proctor eyewear.} Proctors carry Tier-0 glasses so they can read keyed materials when candidates raise questions.
\item[R7] \textbf{Contingency.} Spare eyewear in a range of prescriptions and a supervised Tier-B room are available in every centre.
\item[R8] \textbf{Layering.} PTP adds to existing possession controls and to content-level measures such as SEOYUN; it does not replace them.
\end{enumerate}

\section{Universal Design for Learning}
\label{sec:udl}
A protocol that makes everyone wear special glasses could easily become a new barrier. PTP therefore takes UDL as a design constraint from the start rather than adding accommodations afterwards~\cite{rose2002teaching,cast2024udl,thompson2002universal}. Table~\ref{tab:udl} maps the protocol to the three UDL principles, which UDL~3.0 phrases as \emph{design multiple means} of engagement, representation, and action and expression~\cite{cast2024udl}.

\begin{table}[t]
\centering\small
\begin{tabularx}{\linewidth}{@{}p{24mm}XX@{}}
\toprule
UDL principle & PTP feature & Constraint it imposes \\
\midrule
Representation (perception) & Same item content in four tiers: keyed print, keyed print with Rx, enhanced MR view with speech, braille/audio/large print & Item equivalence across tiers must be verified, as for existing alternate formats \\
Representation (customization) & Tier-A magnification, contrast and polarity remapping, reading guide & Settings chosen by the candidate in advance and rehearsed, never changed by others mid-session \\
Action and expression & Unkeyed answer sheets and scratch paper; existing assistive answer modes unchanged & Keying must not reach the response channel \\
Engagement & Practice materials and loaner Tier-0 glasses before the examination & Familiarization is equally available to all candidates \\
\bottomrule
\end{tabularx}
\caption{Mapping of PTP to the UDL principles~\cite{cast2024udl}.}
\label{tab:udl}
\end{table}

\paragraph{The glasses as a shared assistive platform}
Requiring eyewear of everyone has an effect beyond security. Today, a candidate who needs a magnifier or a reading device stands out. Under PTP everyone wears authority eyewear, and the difference between Tier 0 and Tier A is a matter of configuration rather than a visible separation. In the language of universal design, an access feature becomes part of the baseline. This benefit is conditional. It holds only if Tier A is comfortable, rehearsed, and never slower than the candidate's usual aid, and the human study must test this rather than assume it (\S\ref{sec:protocol}, E4).

\paragraph{Non-negotiable accessibility constraints}
(a) Access to the questions never depends solely on the optical key (Tier B always exists). (b) The reduced luminance of keyed pages (\S\ref{sec:luminance}) must not fall on low-vision candidates; they may choose Tier A or unkeyed large print instead. (c) Extended time, separate rooms, and other existing accommodations are unaffected. (d) Choice of tier is documented, voluntary within eligibility, and never treated as a signal of suspicion. (e) Disabled candidates and their organizations take part in design and evaluation, not only as research subjects. These constraints follow the legal baseline of the Convention on the Rights of Persons with Disabilities and Korean anti-discrimination law~\cite{un2006crpd,korea2007disability}. Digital Tier-B materials should also conform to WCAG~2.2~\cite{w3c2023wcag22}.

\section{Security Analysis and Failure Modes}
\label{sec:security}
\subsection{What changes for the adversary}
Without PTP, a candidate with consumer AI glasses needs one device that looks like ordinary eyewear. With PTP, the face position is taken. The adversary must conceal a camera at position B or D, attach a circular analyzer to it, keep a line of sight to a page that is itself dim, and pass the existing possession controls and radio survey. PTP does not make this impossible. It removes the cheapest and most natural configuration and requires the adversary to build custom hardware. We make no quantitative claim about how much this deters candidates, and that is a question for the evaluation.

\subsection{Failure modes}
\begin{table}[t]
\centering\small
\begin{tabularx}{\linewidth}{@{}p{30mm}Xp{30mm}@{}}
\toprule
Failure mode & Mechanism & Mitigation and residual risk \\
\midrule
Filtered covert camera & Body or desk camera with a commodity circular analyzer & Possession controls, radio survey, dim-page geometry; residual \\
Camera behind issued lens & Covert camera placed between the analyzer and the eye & Transparent serialized frame, seals, issue at entry; residual for the fit-over route \\
Tampered Tier A & Firmware altered to add radio or answer generation & Authority custody, attestation, wipe; insider risk unchanged \\
Unfiltered leakage & $C_{\text{cam}}\approx2D\sin\Gamma$; structural visibility; recoverable by frame averaging (\S\ref{sec:e2}) & Index-matched planarization; budget $C_{\text{cam}}\sqrt{N_eN_f}<\text{CNR}^*$; measured in E1 \\
Polarization camera & Axis misregistration between glyph and background (\S\ref{sec:sim}) & Fabrication tolerance as acceptance criterion \\
Transcription & Candidate copies items onto unkeyed scratch paper for an unfiltered camera & Slow and conspicuous; proctoring; optional keyed scratch paper \\
Memorization and recall & Content recalled after the session & Out of scope; unchanged from today \\
Pre-session leak & Papers or keying masters leaked & Out of scope; key carries no secrecy anyway \\
Wrong analyzer & Opposite-handedness glasses & Text still legible with inverted polarity; no security effect, only a human-factors one \\
Denial of service & Lighting failure, broken eyewear, faulty print batch & Spares, Tier-B room, print quality control \\
\bottomrule
\end{tabularx}
\caption{Failure modes of PTP. ``Residual'' marks risks that PTP reduces but does not remove.}
\label{tab:failures}
\end{table}

Table~\ref{tab:failures} lists the failure modes. Three deserve emphasis. First, \emph{the key is public} (\S\ref{sec:secrecy}). Any security argument that relies on adversaries lacking an analyzer is wrong. Second, the \emph{fit-over route} reintroduces personal frames on the face and is the weakest point of the eyewear rules. It should be kept small by funding registered prescriptions. Third, the defence against \emph{unfiltered} capture is only as good as the measured leakage. If E1 finds that real phone or glasses cameras recover text from keyed pages, the optical layer fails, and PTP's remaining benefit, displacing personal eyewear, must be justified on its own.

\subsection{Relation to SEOYUN}
PTP and SEOYUN fail under different conditions. SEOYUN fails against non-participating or local solvers. PTP fails against filtered covert cameras. An adversary who defeats PTP with a filtered body camera still faces SEOYUN's content measures, and one who uses a local solver still has to solve PTP's capture problem first. Because the two are layered rather than alternatives, we recommend that they be evaluated together in the physical stage.

\section{Evaluation Protocol}
\label{sec:protocol}
The evaluation proceeds in five stages. A later stage starts only when the stopping criteria of the earlier one are met. All thresholds below are \emph{prespecified decision rules to be fixed before data collection}, not results.

\paragraph{E1: optical bench}
Build K-C1 and K-C2 samples with standard glyph sets. Measure the luminance ratio $\rho$ and the glyph contrast $C$ through Tier-0 analyzers across the visible band, at analyzer rotations of $0^\circ$--$45^\circ$ and at viewing angles of $0^\circ$--$40^\circ$. Capture unfiltered images with a panel of real devices (consumer AI glasses, phones, and a linear-polarization camera) at the distances and angles used in SEOYUN's protocol~\cite{kwon2026seoyun}, with ambient, raking, and flash lighting. The primary metric is the extraction rate: the fraction of items whose text, or answer, is recovered by OCR and by a panel of vision--language models from unfiltered captures. It is compared with plain-paper controls and reported with confidence intervals. Also report the measured $C_{\text{cam}}$ for each device and check it against the photon budget of \S\ref{sec:e2}, using the strongest available reader and multi-frame averaging rather than single frames. Include polarization cameras and samples with deliberate axis misregistration. \emph{Stop rule:} if the unfiltered extraction rate cannot be pushed below a predeclared ceiling, the optical layer is abandoned.

\paragraph{E2: human legibility and comfort}
After ethics approval, run a within-subject comparison of plain paper against keyed paper with Tier 0 at several desk illuminances. Outcomes are reading accuracy, reading speed, a standardized visual-comfort questionnaire, and reported headache or eye strain over a session as long as a real examination. The design is a non-inferiority test against plain paper with a margin declared in advance. Report reports of Haidinger's brushes or other polarization cues.

\paragraph{E3: prescription and fit}
Assess Tier-0-Rx routes 1--3 for visual acuity through the insert, alignment of optical centres, comfort, and logistics. These include the share of candidates served by each route, lead times, and failure and replacement rates in a mock registration cycle.

\paragraph{E4: accessibility co-design}
Co-design Tier A and Tier B with low-vision and blind participants and their organizations. Compare Tier A with each participant's usual aid on accuracy, time, fatigue, and preference. Test item equivalence across tiers. \emph{Stop rule:} Tier A is offered only if it is non-inferior to the usual aid for the participants who choose it.

\paragraph{E5: red team and mock examination}
A red team equipped with commodity analyzers and covert cameras attempts capture under the full rule set (R1--R8) and possession controls, and the results are compared with a control room without PTP. Report success rates, time to first capture, and detections. A full mock examination then tests logistics: issue, reconciliation, lighting, contingencies, and proctor workload.

\paragraph{Reporting}
Preregister hypotheses, margins, and stop rules. Report every device, distance, angle, and lighting condition attempted, including failures. No stage may produce candidate-level suspicion scores (G4).

\section{Discussion}
\label{sec:discussion}
\paragraph{Cost and manufacture}
The largest engineering unknown is producing keyed pages with variable content at examination scale. Patterned retarders have been mass-produced for stereoscopic displays, but they carry fixed patterns. Printing per-item glyph patterns into photo-aligned retarder films~\cite{chigrinov2008photoalignment}, or printing CLC inks of opposite handedness, at the volume of a national examination and under examination security has not, to our knowledge, been demonstrated. Unit costs for keyed booklets and eyewear are therefore left as parameters for a feasibility study rather than estimated here. The computer-based-testing variant may become practical first.

\paragraph{Why compulsory eyewear rather than a better detector}
Detectors compete with miniaturization: each generation of AI glasses looks more like ordinary spectacles. Compulsory issued eyewear does not depend on that race. It does not need to recognize a smart frame, because no personal frame is admitted to the face position at all. The cost is that every candidate must wear something for several hours. That cost is acceptable only if E2 and E4 show non-inferior legibility and comfort, and only if the prescription and accessibility tiers work as specified.

\paragraph{Social acceptability}
The cinema analogy helps because the arrangement is familiar. Examinations nonetheless carry higher stakes, and requiring a device on the face of every candidate needs a legal basis, public consultation, and clear limits: no electronics in Tier 0, no data collection, and no use of eyewear status as evidence of misconduct.

\paragraph{Limitations}
This paper presents analysis and simulation only. The optical model is idealized, the reader in E2-sim is a conventional OCR engine rather than the strongest available reader, and the illustrative parameter values are not measured parts. The key provides no secrecy. Filtered covert cameras remain possible. The fit-over route keeps personal frames on the face. The luminance cost is substantial, and no human data yet show that it is tolerable.

\section{Ethics and Broader Impact}
\label{sec:ethics}
PTP is preventive. It must not be used to flag candidates, infer guilt, or support public allegations. Tier-0 eyewear contains no electronics and cannot collect data. Tier-A devices are assistive tools under the authority's custody, are wiped after each session, and must not be repurposed for monitoring. A broken seal, a request for a different tier, or a need for spare eyewear is a logistical event, not evidence.

Universal eyewear requirements must not fall disproportionately on candidates with disabilities, refractive errors, or sensory sensitivities. The accessibility constraints in \S\ref{sec:udl} are conditions for deployment, not aspirations. Human studies (E2--E5) require ethics approval, informed consent, and the participation of disabled people in design as well as in testing.

The analysis also shows an adversary what PTP does not stop, chiefly filtered covert cameras. We judge that publishing this is better than overstating the protocol. Authorities should not adopt a measure whose limits are hidden from them.

\section{Conclusion}
\label{sec:conclusion}
PEACE-TEST-PROTOCOL brings the cinema's arrangement of glasses at the door into the examination room. The page is encoded in circular polarization and reads as a grey field to the naked eye and to unfiltered cameras. Every candidate wears authority-issued analyzer eyewear, so the face position that AI glasses exploit belongs to the examination authority. Prescription inserts, an offline assistive mixed-reality tier, and non-optical formats make that requirement a platform for universal design rather than a barrier. The protocol offers no cryptographic secrecy and does not defeat a determined adversary with a filtered covert camera. What it changes is the cheapest attack. Simulations support the choice of circular keying and the rejection of spectral keying. They also show that leakage must be budgeted against video averaging rather than single frames. Whether the protocol is worth deploying depends on measurements this paper does not yet have: physical unfiltered leakage, human legibility and comfort, and accessibility outcomes. The evaluation protocol is designed to produce them.

\section*{Author and Availability Statements}
\paragraph{Contributions and tools}
Dohyeon Raphael Kwon conceived the protocol, specifically mandatory issued analyzer eyewear, keyed papers modelled on cinema polarization glasses, prescription inserts on the smart-glasses model, and the universal-design framing, and directed the project. Generative AI tools assisted with manuscript drafting, the optical derivations, simulation code, literature lookup and source verification, and build checks. The author is responsible for the content and for the decision to submit. AI assistance and source checks are not substitutes for independent scientific validation.

\paragraph{Version history}
v1 (21 November 2026): framework, optical analysis, simulation study (E1--E3-sim), eyewear tiers, UDL mapping, failure modes, and evaluation protocol. No physical or human-subject results.

\paragraph{Companion work}
SEOYUN~\cite{kwon2026seoyun} addresses the model stage of the same threat. PTP addresses the capture stage and the eyewear itself. The two papers are designed to be layered.

\paragraph{Affiliation and interests}
The author is the founder of PEACE-UNIVERSE, Cheongju, Republic of Korea (\url{https://peace.ac}). No manufacturer, provider, or examination authority funded or endorsed this work. Product references illustrate existing capabilities and are not recommendations.

\paragraph{Code and data}
The simulation code and generated results are described in Appendix~\ref{app:repro} and will be released with the source archive. Third-party spectral datasets are not redistributed; their sources and checksums are listed. No physical measurements accompany this version.

\bibliographystyle{plainnat}
\bibliography{refs}
\appendix
\section{Derivation of the Unfiltered-Camera Leakage}
\label{app:leak}
Let a page element emit fully circular light with Stokes vector $I_0(1,0,0,\pm1)$: the plus sign for the background and the minus sign for glyphs. Model the camera path as a linear retarder of retardance $\Gamma$ with its fast axis at $0^\circ$, followed by a linear diattenuator with diattenuation $D$ and its axis at $45^\circ$, and finally an intensity sensor. Up to sign convention, the retarder maps $S_3$ into $S_2$ with $S_2'=S_3\sin\Gamma$. The diattenuator then transmits $t\,(S_0+D\,S_2')$. The detected intensities are therefore $I_\pm\propto1\pm D\sin\Gamma$, and the contrast of a glyph against the background is
\begin{equation}
C_{\text{cam}}=\frac{I_+-I_-}{I_+}=\frac{2D\sin\Gamma}{1+D\sin\Gamma}\approx2D\sin\Gamma .
\end{equation}
For other orientations the effect depends on the relative angle between the retarder and diattenuator axes and is largest at $45^\circ$. When $\Gamma=0$ or $D=0$ the contrast vanishes, which shows why intensity and linear-polarization cameras cannot see a circular key without additional retardance.

\section{Illustrative Numbers}
\label{app:numbers}
All numbers in \S\ref{sec:optics} come from the formulas above and the stated assumptions. For the luminance budget, $T_u=0.42$ and $T_a=0.85$ give $\rho=0.42^2\times0.85\approx0.150$ and $1/\rho\approx6.7$. For chromatic crosstalk, with $\lambda_0=550$\,nm, $\delta(450)\approx0.349$\,rad and $\delta(650)\approx-0.242$\,rad, giving $\sin^2(\delta/2)\approx0.030$ and $0.015$. For linear tilt, $1-\tan^2 15^\circ\approx0.928$ and $1-\tan^2 30^\circ\approx0.667$. None of these values is a measurement.

\section{Reproducibility of the Simulation Study}
\label{app:repro}
\paragraph{Software}
The simulations use Python~3.13 with NumPy, SciPy, Pillow, Matplotlib, and \texttt{colour-science}~0.4.7, the last for the CIE 1931 2$^\circ$ observer and D65. OCR uses Tesseract~5.5.3 with the \texttt{eng} and \texttt{kor} models. The random seeds are fixed in the scripts: 0 for E1, 20261120 for E2, and 7 for E3.

\paragraph{Data}
Camera spectral sensitivities for 28 cameras come from~\cite{jiang2013space}, retrieved via Zenodo record 3245883. The MD5 of \texttt{camspec\_database.txt} is \texttt{dcf7ce9ec9f2c87b87014f5cc46f8c15}, matching the published checksum. Individual colour-matching functions for 151 observers come from~\cite{asano2016individual}, retrieved via Zenodo record 3252742. The MD5 of \texttt{Data\_151Obs.xls} is \texttt{1c8ea892045faa785ca29c80b7f7674e}, matching the published checksum. Both are used at 400--700\,nm in 10\,nm steps.

\paragraph{Pipeline}
\texttt{exp1\_page\_optics.py}, \texttt{exp2\_capture\_ocr.py}, and \texttt{exp3\_metamerism.py} write JSON logs. \texttt{make\_results.py} turns these into the \LaTeX{} macros and tables used in \S\ref{sec:sim}, so no number in that section is typed by hand.

\end{document}